\documentclass{Interspeech2024}

\interspeechcameraready

\usepackage[utf8]{inputenc}
\usepackage{times}
\usepackage{latexsym}
\usepackage{graphicx}
\usepackage{subcaption}
\usepackage{array}
\usepackage{colortbl}
\usepackage{booktabs}
\usepackage{tabularx}
\usepackage{subcaption}
\usepackage{multirow}
\usepackage{amsmath}
\usepackage{amssymb}
\usepackage[T1]{fontenc}
\usepackage[utf8]{inputenc}
\usepackage{microtype}
\usepackage{inconsolata}
\usepackage{xcolor}
\title{Spoken Word2Vec: Learning Skipgram Embeddings from Speech}

\name[]{Mohammad Amaan}{Sayeed}
\name[]{Hanan}{Aldarmaki}

\address{
  Mohamed Bin Zayed University of Artificial Intelligence
  }
\email{mohammad.sayeed@mbzuai.ac.ae, hanan.aldarmaki@mbzuai.ac.ae}


\begin{document}

\maketitle

\begin{abstract}
Text word embeddings that encode distributional semantics work by modeling contextual similarities of frequently occurring words. Acoustic word embeddings, on the other hand, typically encode low-level phonetic similarities. Semantic embeddings for spoken words have been previously explored using analogous algorithms to Word2Vec, but the resulting vectors still mainly encoded phonetic rather than semantic features. In this paper, we examine the assumptions and architectures used in previous works and show experimentally how shallow skipgram-like algorithms fail to encode distributional semantics when the input units are acoustically correlated. We illustrate the potential of an alternative deep end-to-end variant of the model and examine the effects on the resulting  embeddings, showing positive results of semantic relatedness in the embedding space. 
\end{abstract}

\section{Introduction}

Word embeddings are vector representations that encode distributional relatedness among words, which is correlated with semantic relatedness \cite{mikolov2013linguistic, finley2017analogies, aldarmaki2018unsupervised}. Word2Vec algorithms, like skipgram with negative sampling (SGNG) \cite{mikolov2013distributed} and continuous bag-of-words (CBOW) \cite{mikolov2013efficient}, have been influential in shaping recent progress in statistical NLP, enabling generalization and transfer across tasks and languages, and paving the way to more robust and efficient language models. Whereas text consists of discrete and repeating units, speech signals are composed of continuous and variable signals that need to be processed in order to identify repeating units (e.g. phones/syllables) and word boundaries. Mel Frequency Cepstral Coefficients (MFCCs) or self-supervised neural acoustic models like Wav2Vec \cite{baevski2020wav2vec} are commonly used to extract sequences of acoustic features from continuous speech signals to help distinguish phonetic classes, but they also encode non-linguistic features such as speaker characteristics and environmental acoustics. Supervised learning with large training data that include multiple speakers and noise conditions can lead to intermediate representations that better encode phonetic features and discard irrelevant sources of variation. However, unsupervised and transfer learning are more suitable for low-resource languages, and 
given the success of unsupervised word embeddings in the text modality, attempts at unsupervised acoustic word embeddings have been recently proposed \cite{levin2013fixed, chung2016audio, ghannay2016acoustic, yuan2018learning, peng2020correspondence, kamper2020multilingual}. 
These acoustic embeddings encode the phonetic content of each segment (e.g. word segment) and can help in textless NLP applications such as spoken term discovery and query by example search \cite{aldarmaki2022unsupervised}; they could also potentially be used for unsupervised translation as demonstrated in text \cite{aldarmaki2018unsupervised}. 
While phonetic embeddings arise more naturally in speech models, there have been a few recent attempts at unsupervised semantic embeddings from speech that are meant to encode semantic relatedness in the manner of their text-based counterparts. 
The first such attempt is Speech2Vec \cite{chung2018speech2vec}, which follows the Word2Vec context prediction paradigm, but using a sequential encoder-decoder network to process and reconstruct acoustic features of context words. However, the reported results could not be replicated in subsequent studies and were shown to be flawed in \cite{chen2023reality}\footnote{Our own attempts to replicate the Speech2Vec model did not result in semantic embeddings either.}.
Another attempt was proposed in \cite{chen2019phoneticandsemantic}, where they construct semantic embeddings from speech in two stages: first, they create phonetic embeddings from the acoustic features of each word segment, then use the resulting frozen embeddings as input to a skipgram-like model. While some semantic similarities were observed, 
the results demonstrate mostly phonetic rather than semantic features, demonstrating the difficulty of disentangling semantic from phonetic content of spoken words.  

In these models, the words are pre-segmented by force-alignment using transcribed data, and the models are evaluated by grouping words of the same type together. These simplifications are employed to minimize the many sources of ambiguities in speech that further complicate the development and evaluation of semantic embedding. Yet, no existing model provides satisfactory results comparable to text-based embeddings.    
In this work, we examine the capabilities of architectures employed in previous works in modeling semantic relatedness from speech units. Our results indicate that even with simplifying assumptions, a two-stage approach \`a la \cite{chen2019phoneticandsemantic} is incapable of learning semantic embeddings since the input units are phonetically correlated. We propose an end-to-end and deep variant of the model that shows far more promising results. The main contributions of this paper can be summarized as follows:

\begin{enumerate}

\item We experimentally examine the two-stage architecture proposed in previous work in idealized conditions, and propose an end-to-end variant of the model. We find that a two-stage architecture is incapable of learning distributional features due to the phonetic correlations of the input units, while an end-to-end architecture with sufficient scale and depth can potentially learn distributional semantic features from suitable discretized units. 
\item We apply frame-level clustering using MFCC, Wav2Vec2.0, and HuBERT as acoustic features,
and analyze the resulting semantic embeddings from the solution outlined above.  Our results underline the importance of obtaining discrete and coherent units to move beyond low-level acoustics to corpus-level semantics\footnote{All scripts for replicating our experiments can be found at \\\url{github.com/rainmaker29/SpokenWord2Vec}}. 
\end{enumerate}

\section{Related Work}
\subsection{Word embeddings and subword units}

Skip-gram with negative sampling (SGNS) \cite{mikolov2013distributed} is a canonical example of context-predicting word embedding models, which are obtained by training a network to predict neighboring words and have been shown to be superior to the count-based word vectors previously used for text classification \cite{baroni2014don}. Context prediction is a simple objective that doesn't require more than raw text for training, and it enables unsupervised and efficient encoding of word distributional semantic relatedness by mapping words that appear in similar contexts into neighboring points in a vector space. SGNS is optimized by distinguishing a positive example from \textit{k} negative examples using logistic regression, where the positive examples are neighboring context words and negative examples are randomly drawn from the corpus. Clearly, the model relies on the concept of a coherent and distinct word that can be observed and identified repeatedly and modeled independently from  other words, which is achieved using one-hot vectors to uniquely encode each word in the vocabulary. Subword models like FastText \cite{bojanowski2017enriching} which composes word embeddings from character n-grams,  still rely on coherent words and character n-grams as unique and repeated units. Character-based language models have been successfully implemented, but they require larger networks to model long-term dependencies and some reliance on words or n-grams for the training objective (e.g. consider the modifications for character LMs in \cite{bojanowski2015alternative}, the softmax over words in ELMo \cite{peters-etal-2018-deep}, and Byte-Pair-Encoding or other subword tokenization schemes used in recent large language models \cite{bostrom2020byte}). 

In the speech domain, recent works explored the potential of unsupervised speech recognition \cite{baevski2021unsupervised} and spoken language models \cite{lakhotia2021generative,borsos2023audiolm}. These models employ clustering of the acoustic features to obtain discrete units: for example, \cite{baevski2021unsupervised} applied K-means clustering of Wav2Vec2.0 frames for phonetic segmentation before mean-pooling and dimensionality reduction. \cite{borsos2023audiolm} clustered the acoustic features using K-Means and used the centroid cluster IDs as semantic tokens. HuBERT \cite{hsu2021hubert} is a self-supervised acoustic model that relies on clustering for the discrete target units and has been shown to be more effective than continuous acoustic models for phonemic discrimination. These results point towards discretization as the most viable option for learning semantic models of speech. 

\subsection{Speech-based word embeddings}
\cite{chung2016audio} describe an ``audio version of Word2Vec", which is essentially a denoising sequential auto-encoder using MFCC features as inputs representing pre-segmented words. Given the nature of the model and training objective, the resulting embeddings simply compress the input features and efficiently encode the phonetic content of the spoken words. Speech2Vec was proposed next \cite{chung2018speech2vec} with an objective that more closely resembles Word2Vec models; it uses a sequential encoder-decoder network to reconstruct the acoustic features (i.e. MFCCs) of neighboring words. The original paper reports positive results with good performance in several word similarity benchmarks. The resulting pre-trained Speech2Vec embeddings were released\footnote{\url{https://github.com/iamyuanchung/speech2vec-pretrained-vectors}}, but the code to re-train the embeddings was not. In a recent study, \cite{chen2023reality} examined these pre-trained embeddings by analyzing their properties and vocabulary content, and concluded that they are most likely produced by a text-based model. For example, the released embeddings show little similarity between homophones (words with identical pronunciation like 'hail' and 'hale').
Our attempts to replicate the Speech2Vec model confirm that the resulting speech embeddings encode phonetic rather than semantic similarity, more or less similar to previously proposed phonetic embeddings. While the objective function seems logically similar to the Continuous Bag-of-Words model (CBOW), we surmise that the reconstruction of the phonetic features as the target objective diverts the model from the high-level semantic task to encode low-level phonetic features instead.  \cite{chen2019phoneticandsemantic} proposed a two-stage model that consists of a phonetic module, followed by a linear skipgram-like embedding module. The phonetic module consists of a phonetic encoder and speaker encoder trained in tandem, where the phonetic encoder is optimized by maximizing reconstruction loss and minimizing speaker discriminability. After training the phonetic module,  the resulting phonetic embeddings are frozen and used as input in the second stage. The second module is similar to the standard SGNS model except that the input words are given as phonetic embeddings instead of one-hot vectors. Examining the resulting embeddings, we see that they mostly encode phonetic rather than semantic features.

\section{Architectures \& Scale: Sanity Check}\label{sec:scale}

In this section, we attempt to evaluate the potential of the model described in  \cite{chen2019phoneticandsemantic} for encoding semantic features in a more simplified setting and with various scales.
We examine the architecture in an idealized setting where we use character sequences to represent words instead of acoustic features. Characters are coarsely related to phonemes, but are easier to work with in the following ways: while phones vary across time and frequency domains and can be difficult to segment and identify, each character can be uniquely and perfectly identified and encoded independently from other characters using one-hot vectors. This means that any difficulty we face with character inputs is likely to be amplified with acoustic inputs. On the other hand, positive outcomes with characters will not necessarily generalize to acoustic inputs, but may point to possible fruitful directions for future research. 
\begin{figure*}[ht]
  \centering
  \begin{subfigure}{0.3\textwidth}
    \centering
    \includegraphics[width=\textwidth]{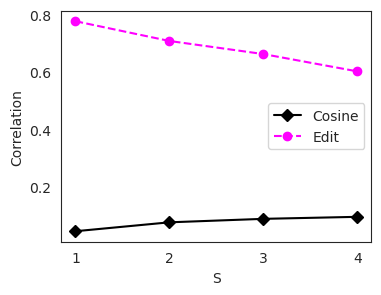}
    \caption{}
    \label{fig:twostage}
  \end{subfigure} 
  \begin{subfigure}{0.3\textwidth}
    \centering
    \includegraphics[width=\textwidth]{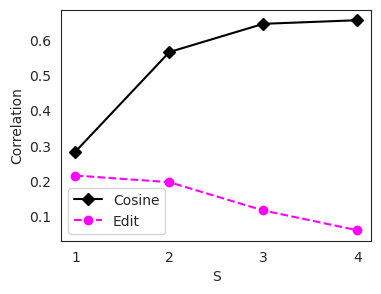}
   \caption{}
    \label{fig:end2end}
  \end{subfigure}
  \begin{subfigure}{0.3\textwidth}
    \centering
  \includegraphics[width=\textwidth]{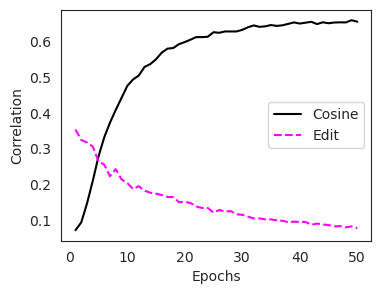}
    \caption{}
    \end{subfigure}
   \caption{Pearson correlation with target cosine distance and edit distance for different scales (s) of the (a) two-stage model vs. (b) end-to-end model. (c) shows the progress of correlation scores throughout the first 50 epochs of training the end-to-en model with s=4.}
  \label{fig:text_figures}
\end{figure*}

\begin{figure}[h]

\begin{subfigure}{.5\textwidth}
\centering
\includegraphics[width=0.9\columnwidth]{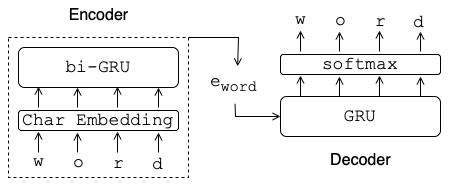}
\caption{encoder-decoder networks}
\end{subfigure}
\begin{subfigure}{.5\textwidth}
\centering
\includegraphics[width=0.95\columnwidth]{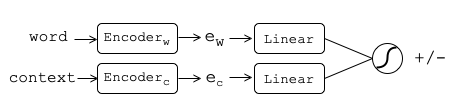}
\caption{Modified skip-gram model}
\end{subfigure}
\caption{Character-based skip-gram model consisting of GRU character encoder, followed by a multi-layer linear network to handle continuous input. The encoder can be pre-trained using a reconstruction loss with the GRU decoder as in (a), or end-to-end within (b). }
\label{fig:skipgram}
\end{figure}

\subsection{Models}

To mimic the phonetic embeddings used in the first stage of \cite{chen2019phoneticandsemantic}, we train a bidirectional GRU auto-encoder with reconstruction loss using character sequences of individual words. The resulting embeddings encode orthographic similarities between words. In the second stage, we freeze these orthographic embeddings and use them as input to a model similar to SGNS, which encodes the input words via linear projection layers, followed by logistic regression to distinguish positive from negative pairs (see Figure \ref{fig:skipgram}). The second stage is identical to the original SGNS model, but since the input is continuous, we use multiple layers of linear projection.  In another setting, we train the model end-to-end with the same SGNS objective: the character sequences are first embedded into fixed-dimensional vectors using the encoder network, followed by linear projection and logistic regression. For each positive context word,  we randomly sample 10 negative words from the corpus. 

\subsection{Data, training, and evaluation scheme}\label{sec:eval}
We use the text transcriptions of the train-clean-100 subset of LibriSpeech \cite{panayotov2015librispeech} for training. We start with the following hyper-parameters: GRU layers = 1,  hidden units = 50, and linear projection layers = 1. We scale the model by multiplying these values by the scaling factor \texttt{s} to observe the effect of scale on performance, with \texttt{s} ranging from 1 to 4 (e.g. with \texttt{s}=4, we have 4 GRU layers, 4 linear layers, and 200 hidden units). For skipgram settings, we used a window of size 5 for the positive examples, and 10 random negative samples without sub-sampling. All models were trained with the Adam optimizer for 100 epochs. To evaluate the resulting embeddings, we train regular SGNS embeddings using the \texttt{gensim} package \cite{rehurek2011gensim} on the same text, then calculate  pairwise cosine distances of a set of word pairs\footnote{We extract a total of 40,000 word pairs based on both semantic and phonetic similarity using a script that will be shared.}, and report the Pearson correlation coefficient between the distances generated by our models and the target distances, as well as the correlation with Lavenshtein edit distance\footnote{We use the Lavenshtein python package: \url{https://pypi.org/project/Levenshtein/}}. A high correlation with edit distance indicates that the embeddings encode orthographic (and hence phonetic) features, whereas a high correlation with \texttt{gensim} cosine distances indicate that the model encodes the desired semantic features.  

\subsection{Results}

\begin{figure*}[ht]
  \centering
  \begin{subfigure}{0.3\textwidth}
    \centering
    \includegraphics[width=\textwidth]{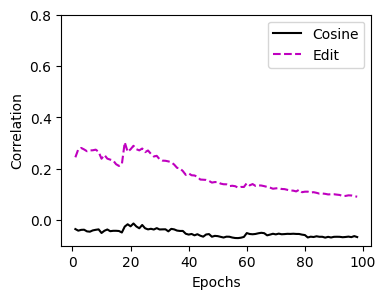}
    \caption{}
    \label{fig:mfcc_cid}
  \end{subfigure} 
  \begin{subfigure}{0.3\textwidth}
    \centering
    \includegraphics[width=\textwidth]{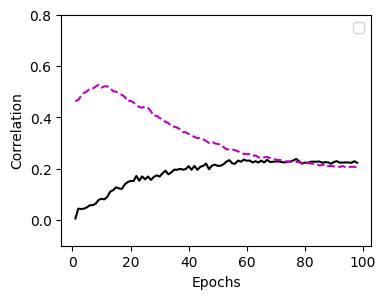}
   \caption{}
    \label{fig:w2v_cid}
  \end{subfigure}
  \begin{subfigure}{0.3\textwidth}
    \centering
  \includegraphics[width=\textwidth]{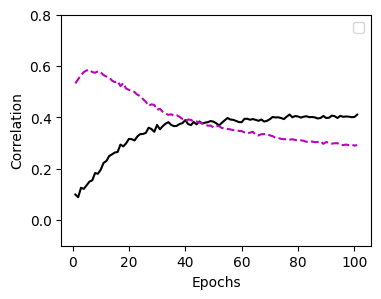}
    \caption{}
    \label{fig:hubert_cid}
    \end{subfigure}
   \caption{Pearson correlation with target cosine distances and edit distances across the 100 training epochs of the end-to-end model with s=4. The input features are obtained by clustering acoustic frames with K=100 using (a) MFCC, (b) Wav2Vec2, and (c) HuBERT features.}
  \label{fig:corr_cluster_figures}
\end{figure*}

The results are shown in Figure \ref{fig:text_figures} (a) and (b).
The final embeddings of a two-stage model (Figure \ref{fig:text_figures} (a)) are highly correlated with edit distance, with almost no correlation with the target \texttt{gensim} embeddings, and we observe little improvement with scale. We conclude that such a two-stage architecture is not suitable for learning semantic embeddings from subword units (e.g. phonetic features) since the frozen embeddings are continuous and highly correlated by their orthographic/phonetic content, which will necessarily be carried forward to the second stage. Training the model end-to-end, on the other hand ((Figure \ref{fig:text_figures} (b)), results in much higher correlation with semantic features, and scaling the model results in higher correlation with the target embeddings and lower correlation with edit distance. As seen in Figure \ref{fig:text_figures} (c), which plots the correlation values throughout training epochs for the largest end-to-end model, the model starts with relatively high correlation with edit distances due to the inherent correlations in the input character sequences. However, over time, the model learns to encode the higher-level distributional semantics and ignore low-level similarities. 
We observe a steady increase in semantic correlation and a decrease in edit distance correlation as long as the training loss is decreasing.
\section{Deep skipgram model for spoken words}
The previous results indicate that an end-to-end approach is more suitable for learning skipgram embeddings from low-level features. However, unlike the discrete character sequences, acoustic sequences are continuous and have their own spurious correlations that can affect the learning process
, and may interfere with the ability to learn high-level semantic correlations. To validate these speculations, we trained the same models using acoustic features directly and used them as input to the encoder in place of the character embeddings. In addition, we experimented with frame-level clustering to derive discrete input units. 

\begin{table*}
    \centering
    \setlength{\tabcolsep}{2pt}
    \scalebox{0.8}{
    \begin{tabular}{|c|l|}
        \hline
        \textbf{Model} & \textbf{Examples of nearest neighbors} \\
        \hline
        \multirow{2}{*}{MFCC} & \cellcolor{green!20}{father, further, however, color, friend, thought, rather, tender, pleasant, beginning, clever
} \\
         & \cellcolor{gray!20}{eyes, honest, arm, ominous, irony, eyed, always, life, often, offers, onward
} \\
         & \cellcolor{white!10}{exclaimed, excellent, agreed, observed, assured, innocent, escaped, discovered, horrid, heart, closed
} \\
        \hline
        \multirow{2}{*}{Wav2Vec2} & \cellcolor{green!20}{father, sister, mother, brother, queer, master, daughter, mary, prison, grandmother, mistress
} \\
         & \cellcolor{gray!20}{eyes, arms, arm, ice, face, hands, hangs, tears, boughs, hair, nose
} \\
         & \cellcolor{white!10}{exclaimed, answered, madam, thank, interrupted, asked, dear, remarked, pray, guess, replied
} \\
        \hline
        \multirow{2}{*}{HuBERT} & \cellcolor{green!20}{father, mother, fear, daughter, brother, sister, wife, villefort, friend, husband, grandmother
} \\
        & \cellcolor{gray!20}{eyes, arms, hands, eyelids, shoulders, tears, hairs, knees, face, ice, weariness
} \\
        & \cellcolor{white!10}{exclaimed, yes, asked, responded, replied, cried, murmured, ah, retorted, quoth, answered
} \\
        \hline
    \end{tabular}
    }
    \caption{Examples of nearest neighbors using the acoustic-based embeddings.}
\label{tab:neighbors_kmeans}
\end{table*}

\subsection{Acoustic Features}\label{sec:feats}
We use 39-dimensional MFCCs (the first 13 DCT coefficients plus their delta and acceleration coefficients), which are commonly used as input in previous works on acoustic word embeddings, but they are known to also encode speaker and environmental acoustics. Pre-trained neural acoustic models, such as Wav2Vec 2.0 and HuBERT are known to be more robust against speaker and environmental acoustics. We use the s3prl package \cite{yang21c_interspeech} to extract these features from whole utterances. We then segment the feature sequences according to the corresponding word boundaries obtained by force-alignment\footnote{We use the alignments from: \url{https://github.com/CorentinJ/librispeech-alignments} }. 

\subsection{Discrete Units}\label{sec:discrete}
Using the continuous acoustic feature vectors directly as input to the end-to-end model, the resulting embeddings showed almost zero correlation with the target semantic features. It seems crucial to discretize the input in order to minimize the effect of acoustic correlations. Ideally, the resulting discrete units should consistently correspond to target phonemes, and different occurrences of the same words should have minimum variation. 
To that end, we clustered the acoustic frames using the K-Means algorithm and used the cluster IDs as discrete acoustic units. Following the settings in HuBERT \cite{hsu2021hubert}, we experimented with K=\{100,500\}. We then replaced all contiguous occurrences of the same cluster ID with a single unit to reduce the length of the sequence and minimize redundancy. Furthermore, following the observation in \cite{borsos2023audiolm}, we normalized all features before clustering such that each feature dimension has zero mean and unit variance

\subsection{Results}

We trained the end-to-end model using the discrete units described above, and evaluated the quality of the embeddings by mean-pooling the embeddings of all occurrences of each word in the vocabulary, then calculating the Pearson correlation coefficient between the pair-wise similarities of these embeddings and the target model, in addition to the correlation with edit distance, as described in section \ref{sec:eval} . Figure \ref{fig:corr_cluster_figures} shows the learning curves for the model with the largest scale (s=4) for 100 epochs of training. Using MFCC-based embeddings, the cosine correlation remained around 0 throughout training, while edit distance correlation decreased over time. Wav2Vec2 and Hubert based models show an increasing trend of semantic correlation and a decreasing edit distance correlation over time, which is the desired trend for semantic embedding; the highest correlation with the target cosine distances was 0.41, achieved by the HuBERT-based model, paired with 0.31 correlation with edit distance. Table \ref{tab:neighbors_kmeans} shows examples of nearest neighbors by the end of training. While both Wav2Vec2 and HuBERT-based models show some semantic qualities, the HuBERT-based embeddings appear more consistent. The embeddings show both semantic  (e.g. \textit{eyes} and \textit{tears}) and phonetic qualities (e.g. \textit{eyes} and \textit{ice}).
\section{Discussion \& Conclusions}

In this work, we conducted several experiments to understand the mechanism of learning word distributional semantics from low-level continuous speech features. Our findings underscore the importance of the following factors: (1) the type of acoustic features used as input, (2) the scale and depth of the embedding network, and (3) using end-to-end training paradigm. By combining self-supervised pre-trained acoustic features, discrete frame-level units obtained by clustering these features, and deep embedding networks trained end-to-end with a skipgram-like objective, we obtained word embeddings that exhibit semantic relatedness that goes beyond phonetic similarity. Previous works in this vein failed to obtain such results for several factors: they mainly relied on MFCC features, used shallow networks, and/or failed to account for spurious acoustic correlations in the input units. Our experiments show that each of these factors individually diminish the ability of the model to break free from the inherent acoustic correlations of spoken words and learn corpus-level semantic features. In spite of the positive results we obtained, we acknowledge several difficulties that would be encountered in a more realistic unsupervised setting. In particular, unsupervised word segmentation remains a rather challenging task, so alternative subword segments should be considered and examined for a truly self-supervised semantic embedding model.

\bibliographystyle{IEEEtran}
\bibliography{template}

\end{document}